\documentclass{article}
\usepackage{spconf,amsmath,amsthm,graphicx,bm,amsfonts,amssymb}
\usepackage{cite,url}
\usepackage{algorithm}  
\usepackage{algorithmic}
\usepackage{caption}
\usepackage{epstopdf}
\usepackage[colorlinks]{hyperref}
\usepackage[font = scriptsize]{subfig}
\usepackage{xfrac}
\usepackage[shortlabels]{enumitem}

\usepackage{nicefrac}       
\usepackage{diagbox}
\usepackage[table,xcdraw]{xcolor}
\usepackage{booktabs}
\usepackage{setspace}

\graphicspath{ {figures/} }

\usepackage{xargs}
\usepackage[colorinlistoftodos,prependcaption,textsize=tiny]{todonotes}

\definecolor{DeepSkyBlue}{RGB}{0,191,255}
\newcommandx{\marshall}[2][1=]{\todo[linecolor=DeepSkyBlue,backgroundcolor=DeepSkyBlue!25,bordercolor=DeepSkyBlue,#1]{#2}}

\usepackage[font=small,skip=4pt]{caption}

\newlist{compact_enum}{enumerate}{4}
\setlist[compact_enum, 1]
{label=\arabic{compact_enumi}., 
leftmargin=20.5pt,
labelsep=6.8pt,
parsep=-2.5pt,
topsep=3pt,
}

\title{Robust Nonparametric Distribution Forecast with Backtest-based Bootstrap and Adaptive Residual Selection\vspace{-0.25cm}}

\name{Longshaokan Wang$^\star$, Lingda Wang\thanks{$^1$Work done during internship at Amazon.}$^{1, \dagger}$, Mina Georgieva$^\star$, Paulo Machado$^\star$, Abinaya Ulagappa$^\star$,}
\address{
\\[-5ex] \it Safwan Ahmed$^\star$, Yan Lu$^\star$, Arjun Bakshi$^\star$, Farhad Ghassemi$^\star$ \\
$^{\star}$Amazon \quad $^{\dagger}$ University of Illinois at Urbana-Champaign \\
\texttt{longsha@amazon.com}\vspace{-0.3cm}
}

\begin{document}
%
\maketitle
\begin{abstract}
Distribution forecast can quantify forecast uncertainty and provide various forecast scenarios with their corresponding estimated probabilities. 
Accurate distribution forecast is crucial for planning -- for example when making production capacity or inventory allocation decisions. 
We propose a practical and robust distribution forecast framework that relies on backtest-based bootstrap and adaptive residual selection. 
The proposed approach is robust to the choice of the underlying forecasting model, accounts for uncertainty around the input covariates, 
and relaxes the independence between residuals and covariates assumption. 
It reduces the Absolute Coverage Error by more than $63\%$ compared to the classic bootstrap approaches and 
by $2\%-32\%$ compared to a variety of State-of-the-Art deep learning approaches on in-house product sales data and M4-hourly competition data.
\end{abstract}
\vspace{-0.2cm}

\vspace{-0.3cm}
\section{Introduction}
\vspace{-0.33cm}
\label{sec:intro}
Time series forecasting is crucial in many industrial applications and enables data-driven planning \cite{Larson2001, Hyndman2018, Salinas2019}, 
such as making production capacity or inventory allocation decisions based on demand forecast \cite{Rangapuram2018}. 
Planners or optimization systems that consume the forecast often require the estimated 
distribution of the response variable (referred to as distribution forecast, or DF) instead of only the 
estimated mean/median (referred to as point forecast, or PF) to make informed and nuanced decisions.
An accurate DF method should ideally 
factor in different sources of forecast uncertainty, including uncertainty associated 
with parameter estimates and model misspecification \cite{Hyndman2018}. 
Furthermore, when deploying a DF method in industrial applications, there are other important practical considerations 
such as ease of adoption, latency, interpretability, and robustness to model misspecification. 
To this end, we propose a practical and robust DF framework that uses backtesting \cite{Bailey2016}
to build a collection of predictive residuals and an adaptive residual selector to pick the relevant residuals for bootstrapping DF.
The proposed framework incorporates different sources of forecast uncertainty by design, 
integrates well with an arbitrary PF model to produce DF, is robust to model misspecification, 
has negligible inference time latency, retains interpretability for model diagnostics, 
and achieves more accurate coverage than the current State-of-the-Art DF methods in our experiments.

The contributions of this paper are as follows: 
\begin{compact_enum}
\item We propose a flexible plug-and-play framework that can extend an arbitrary PF model to produce DF.
\item We extend the existing approach of bootstrapping predictive residuals \cite{Politis2013, Pan2016} 
by using backtest and covariate sampling to improve efficiency and account for uncertainty in input covariates. 
\item We propose an adaptive residual selector, which relaxes the independence between residuals and covariates assumption and boosts model performance. 
\item We propose a new formula on how bootstrapped residuals are applied during forecasting, which scales the residuals w.r.t. the PF. 
\item Lastly, we empirically evaluate the performance of various DF approaches 
on our in-house product sales data and the M4-hourly \cite{Makridakis2018, Alexandrov2020} competition data. 
The proposed DF approach reduces the Absolute Coverage Error by more than $63\%$ compared to the classic bootstrap approaches 
and by $2\%-32\%$ compared to a variety of State-of-the-Art deep learning approaches.
\end{compact_enum}

\vspace{-0.45cm}
\section{Related Work}
\vspace{-0.32cm}
\label{sec:related}
The existing approaches for DF include the following categories: 
1. Using models that make parametric distribution assumptions around the response variable and estimate the associated parameters, 
such as state space models \cite{Hyndman2008, Durbin2012, Seeger2016, Rangapuram2018} and estimating model uncertainty 
through posterior distributions in a Bayesian setting \cite{Dunson2005, Reich2010, Yang2012, Yang2016, Gal2016, Zhu2017}; 
2. Using models that explicitly minimize quantile loss and generate quantile forecast, 
such as Quantile Gradient Boosting \cite{Pedregosa2011}, MQ-R(C)NN \cite{Wen2018}, 
and Temporal Fusion Transformers \cite{Lim2020}; 
and 3. Using variations of bootstrap methods that sample residuals to generate multiple bootstrap forecasts 
and then compute sample quantiles of the bootstrap forecasts \cite{Berkowitz2000, Rao2012, Politis2013, Keogh2013, Pan2016, Kuffner2017}. 
The bootstrap methods have the practical advantage of being able to integrate with an arbitrary PF model to obtain DF, 
but the classic bootstrap methods are usually designed under strong assumptions around the PF model and dataset, 
which might not work well with complex real-world data or modern machine learning PF models. 
The ``delete-$\mathbf{x}^t$'' approach of bootstrapping predictive residuals \cite{Politis2013, Pan2016} by design 
has less assumptions on the data and is more robust to the choice of PF model; 
however, it only computes one predictive residual for each model training, ignores the uncertainty in covariates, 
and assumes that the residuals are independent from the covariates and PF. 
Our proposed DF framework is a significant generalization of the ``delete-$\mathbf{x}^t$'' approach 
and addresses each of the aforementioned limitations.
Another closely related approach is the Level Set Forecaster \cite{Hasson2021}, which is similar to the special case of our approach 
where the in-sample forecast is used for computing and selecting residuals.

\vspace{-0.45cm}
\section{Method}
\begin{figure*}[t]
  \centering
  \includegraphics[width=1.7\columnwidth]{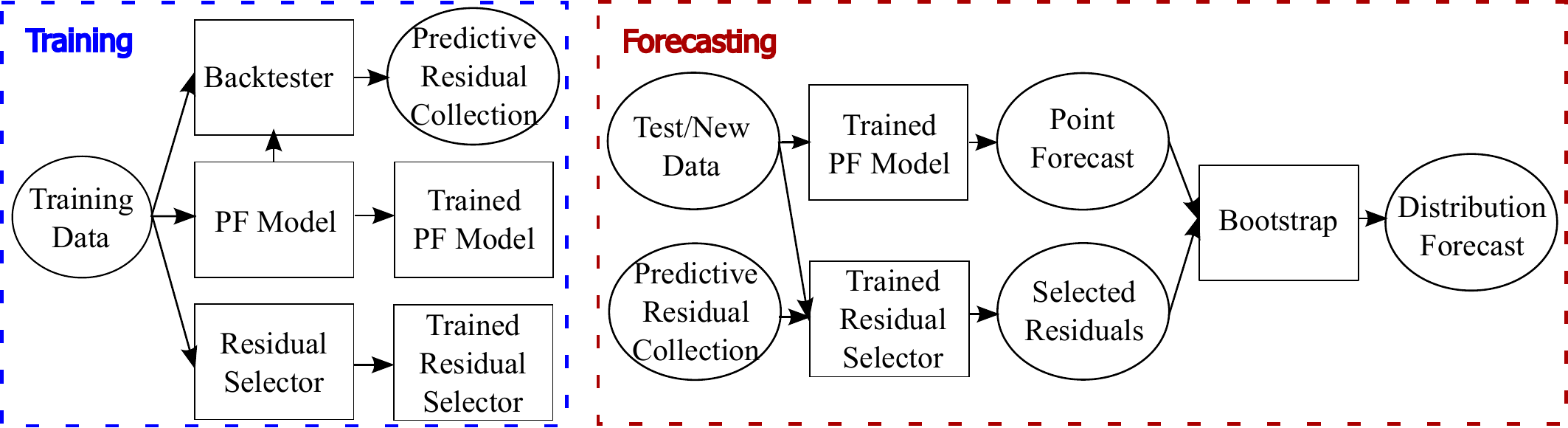}
  \caption{Overview of the proposed DF framework. 
  The backtester generates a collection of predictive residuals; 
  the residual selector selects a subset of residuals for each future data point; 
  the bootstrapping step combines the PF and selected residuals to generate DF.}
  \label{fig:overview}
  \vspace*{-5mm}
\end{figure*}

\begin{figure}[t]
  \centering
  \includegraphics[width=1.0\columnwidth]{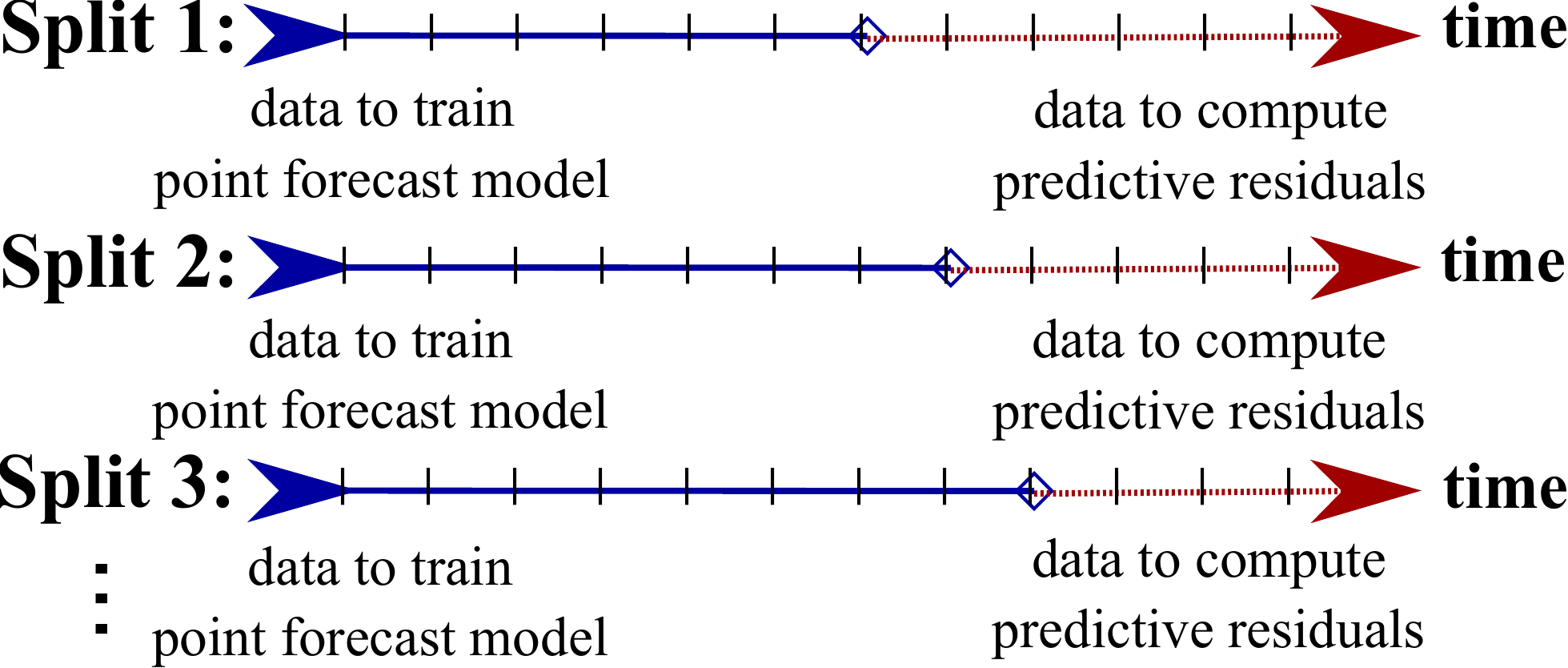}
  \caption{Illustration of building a collection of predictive residuals with backtest. 
  The training split is used to train the PF model, and the test split is used to compute the predictive residuals.}
  \label{fig:bt}
  \vspace*{-5mm}
\end{figure}

\vspace{-0.33cm}
\label{sec:method}
The proposed DF framework is composed of a backtester, a residual selector, and a PF model (Figure \ref{fig:overview}). 
To summarize how it works: During training: 1. Backtest \cite{Bailey2016} is performed on the training data with the PF model 
to build a collection of predictive residuals (Figure \ref{fig:bt}); 
for covariates that need to be estimated for future time points 
(e.g., future price of a product), their values can be sampled from estimated distributions 
during backtest to account for the uncertainty in covariates.
2. The residual selector is pre-specified or learned from the training data as a set of rules or separate machine learning model 
that selects the most relevant subset of predictive residuals given a future data point based on their meta information.
3. Lastly, the PF model is trained on the entire training data. 
During forecasting: 1. For each future data point of interest, the trained PF model generates the PF. 
2. The residual selector selects a subset of residuals. 3. Lastly, random samples of residuals are drawn from the subset 
and applied to the PF to obtain multiple bootstrap forecasts that provide the empirical distribution, 
and sample quantiles of the bootstrap forecasts provide the quantile forecasts for arbitrary target quantiles. 
Essentially, we use the empirical distribution of the selected predictive residuals from backtest to estimate 
the distribution of the future predictive residuals and thus the distribution of the future response variable.

\vspace{-0.5cm}
\subsection{Backtesting}
\vspace{-0.15cm}
\label{subsec:backtest}
Let $\mathcal{D} = \{(\mathbf{X}_i^t, Y_i^t)\}_{i=1, 2, \ldots, n}^{t=s_i, s_i + 1, \ldots, d_i}$ be the training data, 
where $\mathbf{X}_i^t$ is the covariates matrix at time $t$, $Y_i^t$ is the response variable at time $t$, 
$s_i$ is the first time point, and $d_i$ is the last time point for time series $i$. 
For nonparametric distribution forecast, it suffices to estimate the conditional quantiles 
$\widehat{Q}_{Y_i^{d_i+k_i}  \, \vline \,Y_i^{s_i:d_i}, \mathbf{X}_i^{s_i:d_i}, \mathbf{X}_i^{(d_i+1):(d_i+k_i)}}(\tau)$ 
for arbitrary target quantile $\tau \in (0, 1)$, where $k_i$ is the number of time points into the future. 
Backtest is essentially a move-forward cross-validation that preserves the order in time for time series data, 
where the test split is always further in time than the training split. 
Let the backtest start time and step size be $a$ and $l$ respectively. 
For each split point $j = a, a+l, a+2l, \ldots, \max_i(d_i)-1$, 
the training data are divided into a training split $\mathcal{A}_j = \{(\mathbf{X}_i^t, Y_i^t) \in \mathcal{D} \, \lvert \, t \leq j\}$ 
and a test split $\mathcal{B}_j = \{(\mathbf{X}_i^t, Y_i^t) \in \mathcal{D} \, \lvert \, t > j\}$; 
the PF model $\widehat{f}_j$ is trained on $\mathcal{A}_j$, 
and predictive residuals are computed as 
$\{Y_i^t - \widehat{f}_j(Y_i^{s_i:j}, \mathbf{X}_i^{s_i:j}, \mathbf{X}_i^{(j+1):t}) \lvert (\mathbf{X}_i^t, Y_i^t) \in \mathcal{B}_j\}$. 
This process generates a collection of predictive residuals $\mathcal{E} = \{\varepsilon_{i,j}^t\}_{i,j,t}$. 
For those covariates that are not available in the future and need to be estimated, 
we can use their historic estimates, sample from their estimated distributions, 
or add simulated noise to create $\widetilde{\mathbf{X}}_i^t$ to replace $\mathbf{X}_i^t$ 
during backtest to account for uncertainty in covariates.

\vspace{-0.47cm}
\subsection{Selecting Residuals}
\vspace{-0.2cm}
\label{subsec:residuals}
Common PF models typically assume that the residuals are i.i.d. and independent from the covariates and the PF itself \cite{Hyndman2018}. 
However, such assumption doesn't always hold in practice for the predictive residuals. 
E.g., the variance of residuals can increase as we forecast further into the future or as the magnitude of PF increases. 
To relax the commonly imposed independence assumption between residuals and covariates
(or more generally any meta information which can include the PF or other variables not in the original covariates), 
an adaptive residual selector can be learned from the training data to select a subset of residuals based 
on the meta information of the predictive residuals from backtest and the future data point, 
$\widehat{g}(\mathcal{E}, \mathcal{M}, \mathcal{M}^\text{future})$,
so that the selected residuals are conditionally i.i.d.. The residual selector should ideally be based on the meta information 
that has a non-negligible impact on the predictive residuals. 
We mention two options for learning the residual selector here: 
1. Compute distance correlation (which can detect both linear and non-linear dependence) 
\cite{Szekely2009, Wang2018} between the predictive residuals from backtest and their corresponding meta information to identify variables 
with the highest distance correlation to the residuals. Then design rules (e.g., set simple thresholds) 
around these variables to select residuals that have a different distribution from 
the distribution of the entire collection of residuals, which can be verified by the Kolmogorov-Smirnov test \cite{Kolmogorov1933}. 
Note that if the residual selector has no impact, the selected residuals should have the same distribution as the entire collection. 
2. Fit a machine learning model, such as a regression decision tree, to predict residuals from their meta information, 
then apply the model to the meta information of future data points to select the corresponding residuals. 
The performance of this model can also be used to check dependence between meta information and residuals -- 
if the residuals are already independent from the meta information pre-selection, then the model should perform poorly. 

\vspace{-0.47cm}
\subsection{Bootstrapping}
\vspace{-0.2cm}
\label{subsec:bootstrap}
We describe two formulae of generating bootstrap forecasts, {\it Backtest-Additive} and {\it Backtest-Multiplicative}. 
They can be applied to either iterative or direct PF models 
(an iterative model recursively consumes the forecast from the previous time point to forecast for the next, 
whereas a direct model generates forecast for a future time point directly from covariates \cite{Lim2021}).
For Backtest-Additive, to generate bootstrap forecasts for the next time point $d_i + 1$, after obtaining the PF 
$\widehat{Y}_i^{d_i+1} = \widehat{f}(Y_i^{s_i:d_i}, \mathbf{X}_i^{s_i:d_i}, \mathbf{X}_i^{d_i+1})$
and the selected predictive residuals from backtest $\mathcal{G} = \widehat{g}(\mathcal{E}, \mathcal{M}, \mathcal{M}_i^{d_i+1})$,
random samples are drawn from the selected residuals $\varepsilon_b \in \mathcal{G}$ for $b=1, 2, \ldots, B$, 
then the bootstrap forecasts are given by 
$\widehat{Y}_{i, b, \text{Add.}}^{d_i+1} = \widehat{Y}_i^{d_i+1} + \varepsilon_b$. 
Quantile forecasts are obtained by taking sample quantiles of the bootstrap forecasts. 
Generalizing to arbitrary future time point $d_i + k_i$, for an iterative PF model, 
bootstrap forecasts are recursively generated for the next time point until $d_i + k_i$; 
for a direct PF model, the calculation remains the same as 1-step forecast with $d_i+1$ replaced by $d_i+k_i$. 
Note that for a direct PF model, quantile forecasts can be obtained by skipping the residual 
sampling step and adding the sample quantiles of the selected residuals to the PF. 
The formula for Backtest-Additive is similar to the existing approach to bootstrapping predictive residuals \cite{Politis2013, Pan2016}
(while the backtest and residual selection steps are novel in Backtest-Additive). 
The performance of Backtest-Additive can degrade if the variance of residuals increases with the magnitude of the PF, 
or if the magnitude of the future PF is very different from the magnitude of the response variable seen during backtest. 
Hence we also propose Backtest-Multiplicative, which scales the residuals based on the PF: 
After obtaining the PF and the selected residuals in the same way as Backtest-Additive, 
the error ratios are computed by dividing the extracted residuals over their corresponding forecast (or response variable) during backtest, 
$\mathcal{R} = \{\nicefrac{\varepsilon_{h, j}^t}{\widehat{Y}_{h, j}^t} \, \lvert \, \varepsilon_{h, j}^t \in \mathcal{G}\}$; 
then the bootstrap forecasts for the next time point are given by sampling $r_b \in \mathcal{R}$ and 
$\widehat{Y}_{i, b, \text{Multi.}}^{d_i+1} = \widehat{Y}_i^{d_i+1} \cdot (1 + r_b)$. 
The rest remains the same.

\vspace{-0.35cm}
\subsection{Practical Considerations}
\vspace{-0.2cm}
\label{subsec:prac}
Both the backtest step and the residual selection step can be efficiently parallelized across multiple CPU's/GPU's. 
The backtest step requires multiple model training, 
but it is more efficient than the previous ``delete-$\mathbf{x}^t$'' approach of bootstrapping predictive residuals \cite{Politis2013, Pan2016}
and can be done offline at a lower frequency than updating the PF model. 
The only computational overhead during inference time is the (optional) residual selection given the PF, 
so the additional latency of obtaining DF is negligible. 
Furthermore, once a residual collection from backtest is built, quantile forecast 
for any target quantile can be generated without re-running backtest or retraining the PF model, 
whereas DF methods that explicitly minimize quantile loss typically 
require the target quantile to be specified before model training. 
The backtest-based methods are also relatively interpretable: 
They retain the interpretability of the underlying PF model if the PF model is interpretable; 
even with a less interpretable PF model, one can check the predictive residual distribution 
and model performance on the test split (and model coefficients if applicable) during the backtest step to help identify 
which data points or covariates tend to contribute to large predictive residuals 
and whether the model has systematic bias during out-of-sample forecasting.
The choices of bootstrap formula (Backtest-Additive vs Backtest-Multiplicative), 
denominator of error ratios (backtest forecast vs observed response variable), 
residual selector variation, and PF model can be tuned as hyperparameters. 

\vspace{-0.43cm}
\section{Experiments}
\vspace{-0.32cm}
\label{sec:exp}

We conduct experiments on two real-world time-series datasets: 
an in-house product sales dataset and the M4-hourly competition dataset \cite{Makridakis2018, Alexandrov2020}. 
The product sales dataset consists of daily sales of 76 products between 01/01/2017 and 01/10/2021 
and 147 covariates capturing information on pricing, supply constraints, trend, 
seasonality, special events, and product attributes. 
The standard Absolute Coverage Error (ACE) is used to evaluate the DF performance:
The coverage (CO) of quantile forecast $\widehat{Y}_{i(\tau)}^t$ for target quantile $\tau$ 
over the test set $\mathcal{D}_\text{test}$ is defined as 
$\text{CO}(\mathcal{D}_\text{test}; \tau) = \frac{1}{|\mathcal{D}_\text{test}|}\sum_{\mathcal{D}_\text{test}} I\{Y_i^t \leq \widehat{Y}_{i (\tau)}^t\}$; 
and ACE is defined as 
$\text{ACE}(\mathcal{D}_\text{test}; \tau) = |\text{CO}(\mathcal{D}_\text{test}; \tau) - \tau|$. 
(We also track other metrics such as quantile loss, weighted quantile loss, and weighted prediction interval width; 
the conclusions from different metrics are overall consistent.)
A 100-fold backtest is used for evaluation, which is separate from the backtest used for computing predictive residuals -- 
in each training-test split for evaluation, the latter half of the training split is used to perform a separate backtest 
to build the predictive residual collection without using information from the test split for a fair evaluation.
The reported ACE is averaged across all training-test splits, 24-week forecast horizon for product sales and 48-hour horizon for M4-hourly, 
and the following range of target quantiles: $\tau = 0.1, 0.2, \ldots, 0.9$. 
For experiments with deep learning models, the reported ACE is also averaged across 10 trials due to the fluctuation in model performance.

Compared to other DF approaches, bootstrap approaches have the advantage of extending any PF model to produce DF, 
which makes them easy to adopt and able to potentially retain desired properties of the PF model. 
Thus, the first experiment focuses on comparing the proposed Backtest-Additive (BA) 
and Backtest-Multiplicative (BM) against classic bootstrap approaches for DF: 
bootstrap with fitted residuals (FR) \cite{Hyndman2018} and bootstrap with fitted models (FM) \cite{Berkowitz2000, Pan2016}. 
This experiment is performed on the product sales dataset, as it contains covariates 
which can accommodate the use of standard Machine Learning models as direct PF models. A variety of PF models are used 
to assess the bootstrap approaches' robustness to the choice of PF model, including Ridge Regression \cite{Pedregosa2011}, 
Support Vector Regression (SVR) \cite{Pedregosa2011}, Random Forest (RF) \cite{Pedregosa2011}, 
and Feed-forward Neural Networks (NN) \cite{Pedregosa2011}. 
The proposed bootstrap approaches outperform the classic approaches for all PF models (Table \ref{table:bootstrap}).

The second experiment compares against other State-of-the-Art DF approaches, including Quantile Lasso (QLasso) \cite{Pedregosa2011}, 
Quantile Gradient Boosting (QGB) \cite{Pedregosa2011}, DeepAR \cite{Salinas2019, Alexandrov2020}, 
Deep Factors (DFact) \cite{Wang2019, Alexandrov2020}, MQ-CNN \cite{Wen2018, Alexandrov2020}, 
Deep State Space Models (DSSM) \cite{Rangapuram2018, Alexandrov2020}, and 
Temporal Fusion Transformers (TFT) \cite{Lim2020, Alexandrov2020}.
Because the bootstrap approaches require an underlying PF model, for a fair comparison we use the median forecast 
from each of the aforementioned benchmarks as the PF models to be integrated with the backtest-based bootstrap, 
so they share the same model architecture and hyperparameters. The comparison against QGB and QLasso is performed 
on the product sales data and the comparison against the deep learning models is performed on the M4-hourly data\footnote{
Note on the data: QGB and QLasso are not traditional time series models and 
require engineered covariates available in the product sales data but not in M4-hourly, 
while the deep learning models implemented in GluonTS \cite{Alexandrov2020} package 
require the conditioning and forecast horizons fixed for all time series 
and the product sales data contain time series of varying lengths.}.
The proposed bootstrap approaches integrated with the median forecast 
outperform the default DF from the benchmarks (Table \ref{table:dl}).

The third experiment assesses the robustness of the proposed approaches to model assumptions/hyperparameters. 
DeepAR requires the output distribution to be specified prior to the model learning its parameters. 
In this experiment the backtest-based bootstrap approaches integrated with the median forecast 
are compared against the default DF from DeepAR under a variety of output distribution assumptions on the M4-hourly data. 
The proposed approaches outperform the default DF in 5 out of 6 distribution settings (Table \ref{table:distribution}).

The median or mean forecast from the bootstrap approaches can be 
viewed as the updated PF through Bootstrap Aggregating (Bagging). 
As an ensemble output, the Bagging PF can be potentially more accurate than the original PF. 
The fourth experiment evaluates the relative change in Mean Absolute Percentage Error (MAPE) 
of the Bagging PF compared to the original PF on the product sales data. 
The Bagging PF from the proposed approaches achieves the greatest reduction in MAPE (Table \ref{table:mape}) 
for all PF models; i.e., in addition to providing DF, the proposed approaches can also provide more accurate PF. 
One explanation is that if a PF model has systematic bias during backtest, 
its predictive residual distribution will reflect such bias, so by design the median forecast 
will correct for the bias from backtest when bootstrapping DF (Section \ref{subsec:bootstrap}).

\begin{table}[t!]
\caption{ACE comparison of different bootstrap DF approaches integrated with different PF models.}
\label{table:bootstrap}
\centering
\setlength{\tabcolsep}{2pt}
\resizebox{1.0\columnwidth}{!}{
\vspace*{-2mm}
\begin{tabular}{lcccc}
\toprule
Bootstrap\textbackslash{}PF & Ridge & SVR                           & RF    & NN                            \\
\hline
FR     & $0.102 (-0\%)$ & $0.195 (-0\%)$            & $0.207 (-0\%)$ & $0.176 (-0\%)$                         \\
FM     & $0.095 (-7\%)$ & $0.218 (+12\%)$          & $0.171 (-17\%)$ & $0.125 (-29\%)$                         \\
BA     & $0.069 (-32\%)$ & $0.065 (-67\%)$      & $0.055 (-73\%)$ & $0.077 (-56\%)$ \\
BM    & $\textbf{0.038}  (-\textbf{63}\%)$ & $\textbf{0.061} (-\textbf{69}\%)$ & $\textbf{0.027} (-\textbf{87}\%)$ & $\textbf{0.048} (-\textbf{73}\%)$  \\
\bottomrule
\end{tabular}
}
\vspace*{-3.7mm}
\end{table}

\begin{table}[t!]
\caption{ACE comparison of backtest-based bootstrap integrated with the median forecast vs the default DF.}
\label{table:dl}
\centering
\resizebox{1.0\columnwidth}{!}{
\setlength{\tabcolsep}{2pt}
\vspace*{-2mm}
\begin{tabular}{lcc|ccccc}
\toprule
DF\textbackslash{}Model & QLasso         & QGB            & DeepAR         & DFact   & MQCNN          & DSSM           & TFT            \\ \hline
Default                           & $0.188$          & $0.119$          & $0.102$          & $0.098$          & $0.092$          & $0.136$          & $0.067$          \\
Median + BA                        & $0.114$          & $0.078$          & $\textbf{0.100}$ & $\textbf{0.067}$ & $0.078$ & $0.124$          & $\textbf{0.058}$ \\
Median + BM                        & $\textbf{0.039}$ & $\textbf{0.036}$ & $0.104$          & $0.070$          & $\textbf{0.071}$         & $\textbf{0.112}$ & $0.060$  \\
\bottomrule
\end{tabular}
}
\vspace*{-3.7mm}
\end{table}

\begin{table}[t!]
\caption{ACE comparison of backtest-based bootstrap integrated with the median forecast vs the default DF from DeepAR under different pre-specified output distributions.}
\label{table:distribution}
\centering
\setlength{\tabcolsep}{1.5pt}
\resizebox{1.0\columnwidth}{!}{
\vspace*{-2mm}
\begin{tabular}{lcccccc}
\toprule
DF\textbackslash{}Output Dist. & Neg. Bin.  &  Student's t  & Normal       & Gamma          & Laplace        & Poisson        \\
\hline
Default                                    & $0.102$          & $0.192$          & $0.162$          & $\textbf{0.138}$ & $0.114$          & $0.134$          \\
Median + BA                                 & $\textbf{0.100}$ & $0.169$          & $0.116$          & $0.157$          & $0.094$          & $0.128$          \\
Median + BM                                 & $0.104$          & $\textbf{0.165}$ & $\textbf{0.111}$ & $0.156$          & $\textbf{0.088}$ & $\textbf{0.125}$  \\
\bottomrule
\end{tabular}
}
\vspace*{-3.7mm}
\end{table}

\begin{table}[t!]
\caption{Relative change in MAPE for Bagging PF compared to the original PF.}
\label{table:mape}
\centering
\setlength{\tabcolsep}{2pt}
\resizebox{0.85\columnwidth}{!}{
\vspace*{-2mm}
\begin{tabular}{lcccc}
\toprule
Bootstrap\textbackslash{}PF Model    & Ridge &  SVR & RF     & NN                                                        \\
\hline
FR                      & $+0.8\%$ & $+6.5\%$ & $+0.2\%$ & $+0.7\%$            \\
FM                     & $+0.4\%$ & $+6.6\%$ & $-3.8\%$ & $+2.6\%$             \\
BA                      & $-12.3\%$ & $-21.0\%$ & $-\textbf{10.0}\%$ & $+1.5\%$          \\
BM                      & $-\textbf{22.1}\%$   & $-\textbf{31.8}\%$ & $-5.3\%$  & $-\textbf{13.4}\%$                         \\
\bottomrule
\end{tabular}
}
\vspace*{-6mm}
\end{table}

\vspace{-0.43cm}
\section{Conclusion}
\vspace{-0.32cm}
This paper proposes a robust DF framework with backtest-based bootstrap and adaptive residual selection. 
It can efficiently extend an arbitrary PF model to generate DF, is robust to the choice of model, 
and outperforms a variety of benchmark DF methods on real-world data, 
making the proposed framework well-suited for industrial applications.

\small
\singlespacing
\bibliographystyle{IEEEbib}
\bibliography{icassp}

\begin{thebibliography}{10}

\bibitem{Larson2001}
Paul~D Larson,
\newblock ``Designing and managing the supply chain: concepts, strategies, and
  case studies,''
\newblock {\em Journal of Business Logistics}, vol. 22, no. 1, pp. 259, 2001.

\bibitem{Hyndman2018}
Rob~J Hyndman and George Athanasopoulos,
\newblock {\em Forecasting: principles and practice},
\newblock OTexts: Melbourne, Australia, 2 edition, 2018.

\bibitem{Salinas2019}
David Salinas, Valentin Flunkert, Jan Gasthaus, and Tim Januschowski,
\newblock ``Deep{AR}: Probabilistic forecasting with autoregressive recurrent
  networks,''
\newblock {\em International Journal of Forecasting}, vol. 36, no. 3, pp.
  1181--1191, 2020.

\bibitem{Rangapuram2018}
Syama~Sundar Rangapuram, Matthias~W Seeger, Jan Gasthaus, Lorenzo Stella,
  Yuyang Wang, and Tim Januschowski,
\newblock ``Deep state space models for time series forecasting,''
\newblock in {\em Advances in Neural Information Processing Systems}. 2018,
  vol.~31, Curran Associates, Inc.

\bibitem{Bailey2016}
David~H Bailey, Jonathan Borwein, Marcos Lopez~de Prado, and Qiji~Jim Zhu,
\newblock ``The probability of backtest overfitting,''
\newblock {\em Journal of Computational Finance, forthcoming}, 2016.

\bibitem{Politis2013}
Dimitris~N. Politis,
\newblock ``Model-free model-fitting and predictive distributions,''
\newblock {\em TEST}, vol. 22, no. 2, pp. 183--221, 2013.

\bibitem{Pan2016}
Li~Pan and Dimitris~N Politis,
\newblock ``Bootstrap prediction intervals for linear, nonlinear and
  nonparametric autoregressions,''
\newblock {\em Journal of Statistical Planning and Inference}, vol. 177, pp.
  1--27, 2016.

\bibitem{Makridakis2018}
Spyros Makridakis, Evangelos Spiliotis, and Vassilios Assimakopoulos,
\newblock ``The {M4} competition: Results, findings, conclusion and way
  forward,''
\newblock {\em International Journal of Forecasting}, vol. 34, no. 4, pp.
  802--808, 2018.

\bibitem{Alexandrov2020}
Alexander Alexandrov, Konstantinos Benidis, Michael Bohlke-Schneider, Valentin
  Flunkert, Jan Gasthaus, Tim Januschowski, Danielle~C. Maddix, Syama
  Rangapuram, David Salinas, Jasper Schulz, Lorenzo Stella, Ali~Caner Turkmen,
  and Yuyang Wang,
\newblock ``Gluon{TS}: Probabilistic and neural time series modeling in
  python,''
\newblock {\em Journal of Machine Learning Research}, vol. 21, no. 116, pp.
  1--6, 2020.

\bibitem{Hyndman2008}
Rob Hyndman, Anne~B Koehler, J~Keith Ord, and Ralph~D Snyder,
\newblock {\em Forecasting with exponential smoothing: the state space
  approach},
\newblock Springer Science \& Business Media, 2008.

\bibitem{Durbin2012}
James Durbin and Siem~Jan Koopman,
\newblock {\em Time series analysis by state space methods},
\newblock Oxford university press, 2012.

\bibitem{Seeger2016}
Matthias Seeger, David Salinas, and Valentin Flunkert,
\newblock ``Bayesian intermittent demand forecasting for large inventories,''
\newblock in {\em Proceedings of the 30th International Conference on Neural
  Information Processing Systems}, 2016, pp. 4653--4661.

\bibitem{Dunson2005}
David~B Dunson and Jack~A Taylor,
\newblock ``Approximate bayesian inference for quantiles,''
\newblock {\em Journal of Nonparametric Statistics}, vol. 17, no. 3, pp.
  385--400, 2005.

\bibitem{Reich2010}
Brian~J Reich, Howard~D Bondell, and Huixia~J Wang,
\newblock ``Flexible bayesian quantile regression for independent and clustered
  data,''
\newblock {\em Biostatistics}, vol. 11, no. 2, pp. 337--352, 2010.

\bibitem{Yang2012}
Yunwen Yang and Xuming He,
\newblock ``Bayesian empirical likelihood for quantile regression,''
\newblock {\em The Annals of Statistics}, vol. 40, no. 2, pp. 1102--1131, 2012.

\bibitem{Yang2016}
Yunwen Yang, Huixia~Judy Wang, and Xuming He,
\newblock ``Posterior inference in bayesian quantile regression with asymmetric
  laplace likelihood,''
\newblock {\em International Statistical Review}, vol. 84, no. 3, pp. 327--344,
  2016.

\bibitem{Gal2016}
Yarin Gal and Zoubin Ghahramani,
\newblock ``Dropout as a bayesian approximation: Representing model uncertainty
  in deep learning,''
\newblock in {\em Proceedings of the 33rd International Conference on Machine
  Learning}. 2016, ICML'16, pp. 1050--1059, JMLR.org.

\bibitem{Zhu2017}
Lingxue Zhu and Nikolay Laptev,
\newblock ``Deep and confident prediction for time series at {U}ber,''
\newblock in {\em 2017 IEEE International Conference on Data Mining Workshops
  (ICDMW)}. IEEE, 2017, pp. 103--110.

\bibitem{Pedregosa2011}
F.~Pedregosa, G.~Varoquaux, A.~Gramfort, V.~Michel, B.~Thirion, O.~Grisel,
  M.~Blondel, P.~Prettenhofer, R.~Weiss, V.~Dubourg, J.~Vanderplas, A.~Passos,
  D.~Cournapeau, M.~Brucher, M.~Perrot, and E.~Duchesnay,
\newblock ``Scikit-learn: Machine learning in {P}ython,''
\newblock {\em Journal of Machine Learning Research}, vol. 12, pp. 2825--2830,
  2011.

\bibitem{Wen2018}
Ruofeng Wen, Kari Torkkola, Balakrishnan Narayanaswamy, and Dhruv Madeka,
\newblock ``A multi-horizon quantile recurrent forecaster,''
\newblock {\em NeurIPS, Time Series Workshop}, 2017.

\bibitem{Lim2020}
Bryan Lim, Sercan~Ö. Arık, Nicolas Loeff, and Tomas Pfister,
\newblock ``Temporal fusion transformers for interpretable multi-horizon time
  series forecasting,''
\newblock {\em International Journal of Forecasting}, vol. 37, no. 4, pp.
  1748--1764, 2021.

\bibitem{Berkowitz2000}
Jeremy Berkowitz and Lutz Kilian,
\newblock ``Recent developments in bootstrapping time series,''
\newblock {\em Econometric Reviews}, vol. 19, no. 1, pp. 1--48, 2000.

\bibitem{Rao2012}
Tata~Subba Rao, Suhasini~Subba Rao, and C.R. Rao,
\newblock {\em Time Series Analysis: Methods and Applications},
\newblock North Holland, 1 edition, 2012.

\bibitem{Keogh2013}
Gerard Keogh,
\newblock ``The splice bootstrap,''
\newblock {\em arXiv preprint arXiv:1311.5828}, 2013.

\bibitem{Kuffner2017}
Todd~A. Kuffner, Stephen M.~S. Lee, and G.~Alastair Young,
\newblock ``Optimal hybrid block bootstrap for sample quantiles under weak
  dependence,''
\newblock {\em arXiv preprint arXiv:1710.02537}, 2017.

\bibitem{Hasson2021}
Hilaf Hasson, Yuyang~(Bernie) Wang, Tim Januschowski, and Jan Gasthaus,
\newblock ``Probabilistic forecasting: A level-set approach,''
\newblock {\em NeurIPS}, 2021.

\bibitem{Szekely2009}
Gábor~J. Székely and Maria~L. Rizzo,
\newblock ``{Brownian distance covariance},''
\newblock {\em The Annals of Applied Statistics}, vol. 3, no. 4, pp. 1236 --
  1265, 2009.

\bibitem{Wang2018}
Longshaokan Wang,
\newblock {\em Sufficient Markov Decision Processes},
\newblock Doctoral dissertation, North Carolina State University, 2018.

\bibitem{Kolmogorov1933}
A.~Kolmogorov,
\newblock ``Sulla determinazione empirica di una lgge di distribuzione,''
\newblock {\em Inst. Ital. Attuari, Giorn.}, 1933.

\bibitem{Lim2021}
Bryan Lim and Stefan Zohren,
\newblock ``Time-series forecasting with deep learning: a survey,''
\newblock {\em Philosophical Transactions of the Royal Society A: Mathematical,
  Physical and Engineering Sciences}, vol. 379, no. 2194, 2021.

\bibitem{Wang2019}
Yuyang Wang, Alex Smola, Danielle~C. Maddix, Jan Gasthaus, Dean Foster, and Tim
  Januschowski,
\newblock ``Deep factors for forecasting,''
\newblock {\em ICML}, 2019.

\end{thebibliography}

\end{document}